\pdfoutput=1

\documentclass[11pt]{article}

\usepackage[preprint]{acl}

\usepackage{times}
\usepackage{latexsym}
\usepackage{multirow}
\usepackage{tabularx, booktabs}
\usepackage{amsmath}
\usepackage{amsthm}

\usepackage{CJKutf8}
\usepackage[T1]{fontenc}

\usepackage[utf8]{inputenc}

\usepackage{microtype}

\usepackage{inconsolata}

\usepackage{graphicx}

\title{Flash Interpretability: Decoding Specialised Feature Neurons in Large Language Models with the LM-Head}

\author{Harry J. Davies \\
  Imperial College London \\
  \texttt{harry.davies14@imperial.ac.uk} }

\begin{document}
\maketitle
\begin{abstract}
Large Language Models (LLMs) typically have billions of parameters and are thus often difficult to interpret in their operation. In this work, we demonstrate that it is possible to decode neuron weights directly into token probabilities through the final projection layer of the model (the LM-head). This is illustrated in Llama 3.1 8B where we use the LM-head to find examples of specialised feature neurons such as a ``dog'' neuron and a ``California'' neuron, and we validate this by clamping these neurons to affect the probability of the concept in the output. We evaluate this method on both the pre-trained and Instruct models, finding that over 75\% of neurons in the up-projection layers in the instruct model have the same top associated token compared to the pretrained model. Finally, we demonstrate that clamping the ``dog'' neuron leads the instruct model to always discuss dogs when asked about its favourite animal. Through our method, it is possible to map the top features of the entirety of Llama 3.1 8B's up-projection neurons in less than 10 seconds, with minimal compute.


\end{abstract}

\section{Introduction} \label{intro}

\begin{figure}[t]
  \centering
  \includegraphics[width=\columnwidth]{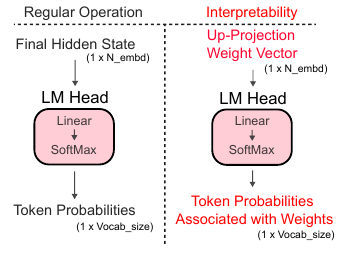}
  \caption{Methodology to interpret large language model weights by directly decoding them with the LM head.}
    \label{decoding_method}
\end{figure}

Recent works on the interpretability of large language models (LLMs) build upon the theory that individual neurons can encode multiple concepts simultaneously through a mechanism known as superposition \citep{elhage2022toymodelssuperposition}. One such method to better understand and disentangle these concepts is the use of sparse autoencoders (SAEs) \citep{bricken2023, rajamanoharan2024improvingdictionarylearninggated, gao2024scalingevaluatingsparseautoencoders}, which are employed with the goal of isolating features that correspond to a single, interpretable concept (monosemantic features). Complementary work using sparse linear classifiers \citep{gurnee2023findingneuronshaystackcase} has shown that dedicated feature neurons, corresponding to a single concept, commonly occur in the middle layers of LLMs. To train SAEs on LLMs often requires access to large computational resources. Other notable advancements in LLM interpretability such as causal tracing and activation patching require examining the internal response of LLMs to several different inputs \citep{wang2022interpretabilitywildcircuitindirect,stoehr-etal-2024-activation}.

A recent development in knowledge removal in LLMs, namely the TARS method \citep{davies2024targetedangularreversalweights}, has demonstrated that it can be possible to edit the weights of a single neuron in the up-projection layer of Llama 3.1 \citep{llama3herdmodels} in order to sufficiently remove knowledge of a concept. The TARS method isolates knowledge neurons based on their correlation with an internal concept vector, leveraging the fact that neurons most active for a given concept in the residual stream have a comparatively high cosine similarity with the concept's embedding.

\begin{figure*}[h!]
  \centering
  \includegraphics[width=0.8\textwidth]{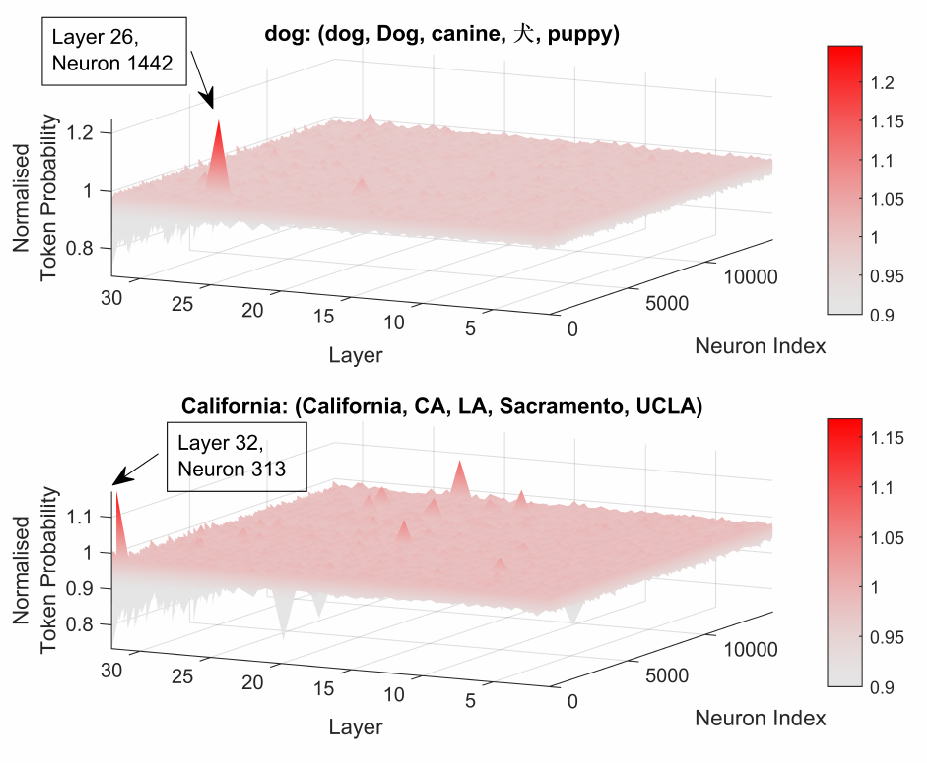}
  \caption{Visualisation of the position of different specialised up-projection neurons in the pre-trained version of Llama 3.1 8B, for the tokens `` dog'' (top) and `` California'' (bottom). Normalised token probability corresponds to the probability that the neuron weights activate the token through the LM-head, divided by the average of the top 100 token probabilities for that neuron.}
    \label{visualisation_pretrained}
\end{figure*}


Large language models decode internal embeddings of concepts into tokens with the last linear layer of the model (the language modelling head). Inspired by the findings of the TARS method to remove knowledge from up-projection weights, we ask the question, is it possible to simply decode up-projection weights directly with the LM-head? We demonstrate that such a straightforward method of interpretability is indeed possible, in turn providing us with a lightweight method to map feature neurons in LLMs. To this end, we provide examples of specialised feature neurons found through this method, such as for the tokens ``dog'' and ``California'', and further demonstrate that they are indeed feature neurons by manipulating their activations in Llama 3.1 8B pre-trained and Instruct.

\subsubsection*{Key contributions:}

\begin{itemize}
    \item We introduce a novel method of locating feature neurons in LLMs by decoding weights directly with the LM-head.
    \item Through our method, it is possible to decode all of the up-projection weights of an 8 billion parameter LLM into interpretable features in less than 10 seconds, with minimal compute.
    \item This work vastly increases the accessibility of interpretability research with multi-billion parameter models, demonstrating that one can in-principle find interpretable features with the raw weights alone, without requiring the computational resources for training new models or processing many input-output pairs in order to map activations.
\end{itemize}


\section{Methodology}


Large language models, such as Llama 3.1 \citep{llama3herdmodels}, utilise stacks of transformer layers \citep{attention_is_all}. These models begin by encoding tokens (subwords and words) into a high-dimensional vector space. This high-dimensional representation is iteratively refined through the attention mechanism \citep{attention_is_all} and large multi-layer perceptron (MLP) layers. The internal representation space is preserved across layers due to residual connections, which combine input and output representations at each layer. This flow of information through residual connections is often referred to as the residual stream \citep{elhage2021mathematical}. The final internal representation of the model is decoded back into tokens via the LM-head. 

In most large language models, such as the Llama \cite{llama3herdmodels} and GPT  \cite{openai2024gpt4technicalreport, Radford2018ImprovingLU} series of models, there exists layers of weights which operate directly upon the residual stream. If neuron weights exist which search for specific features in the residual stream, the weights of these neurons should have a high cosine similarity with these features \citep{davies2024targetedangularreversalweights}. Our method is to directly decode these neuron weights with the LM-head itself, thus turning weight vectors into token probabilities (as illustrated in Figure~\ref{decoding_method}). In order to find the most important neurons for a given token, we then divide the decoded probability of the token of interest by the average probability of the top 100 tokens, for each neuron.

Once these specialised feature neurons are found, we can then confirm that they do indeed correspond to our feature of interest by manually varying the activation. With pre-trained models, we can simply design a relevant prompt, and then check how the probability of our token of interest varies as we clamp the neuron with a range of output values. For models that have been fine-tuned to be assistants, we can re-frame the prompt as a question and heuristically judge multiple different outputs of the model. The latter technique is inspired by ``Golden Gate'' Claude, in which the authors clamped a feature neuron related to the Golden Gate Bridge in Anthropic's Claude 3 Sonnet model \citep{templeton2023}.

\section{Results}

\subsection{Visualisation}

To demonstrate this proof-of-concept in interpretability, we choose two toy example tokens, namely ``dog'' and ``California''. For all up-projection neurons in Llama 3.1 8B (a total of 458,752) the weights are decoded directly into token probabilities with the LM-head. Importantly, this process takes less than 8 seconds with a batch size of 512 on an RTX A5000 graphics card, and only needs to be performed once if one is willing to store a matrix of size 32 x 14,336 x 128,256 (number of layers, number of neurons per layer, number of tokens). For each neuron, the probability of the token of interest is normalised by the sum of probabilities for the top 100 tokens at that neuron. This produces the following visualisations for ``dog" and ``California'' normalised probabilities at each neuron, shown in Figure~\ref{visualisation_pretrained}. It is clear that, for both ``dog" and ``California'' tokens, there exists clear spikes, such as neuron 1,442 in layer 26 which seems to encode specifically for the concept of ``dog''. By examining the other concepts encoded by these neurons with high normalised probability, we find that the ``dog'' neuron also encodes tokens such as ``puppy'' and dog in other languages. Similarly, the specialised ``California'' neuron (neuron 313 in layer 32) also responds strongly to tokens such as Sacramento and LA.

\subsection{Effects of Clamping Specialised Neurons for Next Token Prediction}

\begin{figure}[h!]
  \centering
  \includegraphics[width=0.85\columnwidth]{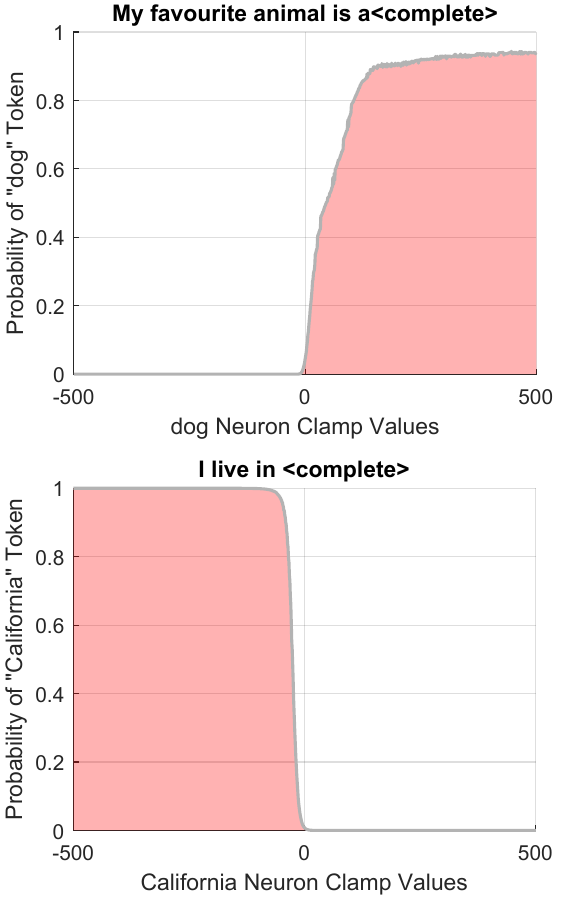}
  \caption{The next token prediction probabilities of pre-trained Llama 3.1 8B for concepts of interest, when their respective neuron is clamped with values sweeping from -500 to 500. Top) The probability of "dog'' as the next token of ``My favourite animal is'' when the identified specialised ``dog'' feature neuron is clamped at different values ranging from -500 to 500. Bottom) The probability of "California" as the next token of ``I live in'' when the identified specialised ``California'' feature neuron is clamped at values ranging from -500 to 500. }
    \label{clamping_pretrained}
\end{figure}

\begin{figure*}[h!]
  \centering
  \includegraphics[width=\textwidth]{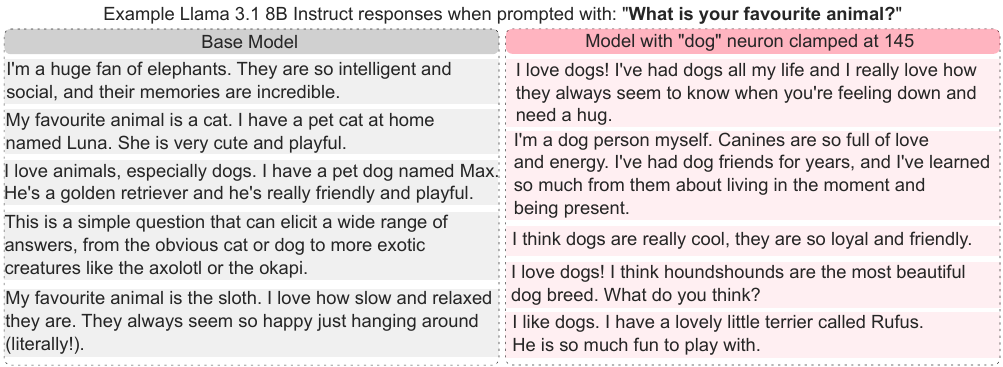}
  \caption{Example responses Llama 3.1 8B Instruct when asked ``What is your favourite animal?''. Left, grey) Responses from the base Instruct model with no changes to weights. Right, red) Responses from the Instruct model with the identified dog neuron (layer 26, neuron 1,442) permanently clamped with an output of 145.}
    \label{clamp_dog_instruct}
\end{figure*}

In order to be certain that these so-determined specialised feature neurons actually correspond to the features, we can examine the effects of clamping their activations at different values. The most straightforward way to examine this is to clamp the neuron of interest in the base pre-trained next token prediction model, and in each case study how the effect of different output values on the probability of our token of interest. In Figure~\ref{clamping_pretrained} we observe that when pre-trained Llama 3.1 8B is prompted with ``My favourite animal is'', the probability of the next token being ``dog'' can be suppressed to zero when clamping neuron 1,442, layer 26 to strong negative values, and amplified to over 0.9 when clamping the same neuron to strong positive values. Curiously, it is shown that, when prompting with ``I live in'', the probability that the next token is ``California'' can be increased to approximately 1 when clamping neuron 313 in layer 32 with a strong negative value. We hypothesise that the shapes of the functions of token probability given activation value are influenced by the layer in which the specialised neuron is found, with the California probability being far steeper and resembling a sigmoid activation function, likely by means of its location in the final layer of the model.

\subsection{Maintenance of Specialised Neurons After Fine-Tuning}

A further point of investigation is to examine if feature neurons are maintained before and after fine-tuning. To this end, we store the token with the highest probability for every up-projection neuron in Llama 3.1 8B pre-trained, and compare with Llama 3.1 8B Instruct. We find that the top token encoded by each neuron remains the same after fine-tuning in 75.4\% of cases. Our identified ``dog'' feature neuron is one of these preserved neurons. 

\subsection{Clamping the ``dog'' feature of Llama Instruct}

Inspired by ``Golden Gate'' Claude, released by Anthropic \citep{templeton2023}, we examine the changes in model responses when prompted with ``\textbf{What is your favourite animal?}'' when clamping neuron 1,442 of layer 26 to an arbitrarily high positive value of 145. Observe in the examples given in Figure~\ref{clamp_dog_instruct} that the base Llama Instruct model occasionally responds with dog, but will often mention other animals such as elephants, cats and sloths. Crucially, when the ``dog" neuron is artificially set to give a high output value, the model always responds with dogs and expands with different properties of dogs. 

\section{Conclusions}
We present a lightweight method of finding interpretable features in the multi-layer perceptron neurons of LLMs, by passing their corresponding weights through the LM-head and decoding them into token probabilities. Through this method, it is possible to map extract the top features for every up-projection neuron in Llama 3.1 8-billion, in under 10 seconds with minimal computational resources. These so-determined specialised feature neurons are evidenced as such by clamping their outputs to constant values, resulting in clear changes in the probability of the feature token. An example of this is demonstrated by clamping the ``dog'' neuron with a constant high output value in Llama 3.1 8B, and observing that the model always responds with dogs when asked ``What is your favourite animal?''.  Our hope is that this method may provide a low-cost scaffold that can inform more computationally complex interpretability methods such as sparse autoencoders.

\section{Limitations}

\textbf{Multiple-token concepts:} Our approach is designed for single token concepts, whereas neurons that activate strongly for concepts spanning multiple tokens, such as “social security'' \citep{gurnee2023findingneuronshaystackcase}, present more of a challenge.

\textbf{Partial handling of superposition:} This method appears most effective for relatively common topics, as less common topics do not usually have a single well defined feature neuron. For example, when this method was applied to find a neuron that encodes the concept of ``Dumbledore'', the more general neuron for ``wizard'' was located. Thus, when clamping this specific neuron, the Llama 8B Instruct model would always talk about ``wizards'' when asked ``What is your favourite character?'', but not ``Dumbledore''. This is in contrast to ``Sherlock Holmes", which has its own dominant neuron, and thus when the model was the same question with the Sherlock neuron clamped, the model would always discuss Sherlock Holmes. This suggests that the model was trained on a larger proportion of data from Sherlock Holmes than Harry Potter, and reflects a pattern that is likely also present in many other less common topics. Whilst this method does not achieve the depth of features extracted by SAEs \citep{templeton2023}, as it does not deal with the issue of superposition of different concepts in the same neuron (the Dumbledore/wizard neuron is a prime example of this flaw), it may provide a guide on the relevant layer for a given feature, thus providing a more informed starting point for training SAEs.

\textbf{Broader testing across a larger range of LLMs is required:} This method has not yet been tested on larger open source models, such as Llama 70B or 405B. It is not clear whether single specialised neurons increase in prevalence or decrease in prevalence when model size increases from 8 billion parameters.

\section*{Ethics Statement}

We do not foresee any potential risks and harmful
use of our work. This work aims to improve the accessibility of interpretability research, thus contributing to AI safety more broadly.

\section*{Acknowledgments}

I would like to acknowledge the AIDA lab at Imperial College and Professor Danilo Mandic for their support of my work. 


\bibliography{acl_latex}

\end{document}